\documentclass[letterpaper]{article} 
\usepackage{aaai25}  
\usepackage{times}  
\usepackage{helvet}  
\usepackage{courier}  
\usepackage[hyphens]{url}  
\usepackage{graphicx} 
\usepackage{natbib}  
\usepackage{caption} 
\usepackage{algorithm}
\usepackage{algorithmic}
\usepackage{amsmath}
\usepackage{booktabs}
\usepackage{makecell}
\usepackage{xcolor}
\usepackage{newfloat}
\usepackage{listings}
\DeclareCaptionStyle{ruled}{labelfont=normalfont,labelsep=colon,strut=off} 
\floatstyle{ruled}
\newfloat{listing}{tb}{lst}{}
\floatname{listing}{Listing}
\title{LightPROF: A Lightweight Reasoning Framework for Large Language Model on Knowledge Graph}
\author{
    Tu Ao\textsuperscript{\rm 1}\equalcontrib, 
    Yanhua Yu\textsuperscript{\rm 1}\protect\thanks{Corresponding author.}\equalcontrib, 
    Yuling Wang\textsuperscript{\rm 2}\protect\thanks{Corresponding author.}, 
    Yang Deng\textsuperscript{\rm 3}, 
    Zirui Guo\textsuperscript{\rm 1}, 
    Liang Pang\textsuperscript{\rm 5}, 
    Pinghui Wang\textsuperscript{\rm 6}, 
    Tat-Seng Chua\textsuperscript{\rm 4}, 
    Xiao Zhang\textsuperscript{\rm 1}, 
    Zhen Cai\textsuperscript{\rm 1}
}
\affiliations{
    \textsuperscript{\rm 1}Beijing University of Posts and Telecommunications, China\\
    \textsuperscript{\rm 2}Hangzhou Dianzi University, China\\
    \textsuperscript{\rm 3}Singapore Management University, Singapore\\
    \textsuperscript{\rm 4}National University of Singapore, Singapore\\
    \textsuperscript{\rm 5}Institute of Computing Technology, Chinese Academy of Sciences, China\\
    \textsuperscript{\rm 6}Xi'an Jiaotong University, China\\

    \{aotu\_bupt, yuyanhua, zrguo, xiao20010420, caizhen\}@bupt.edu.cn,\\
    wangyl0612@hdu.edu.cn,
    pangliang@ict.ac.cn,
    phwang@mail.xjtu.edu.cn,
    dcscts@nus.edu.sg
}

\usepackage{bibentry}

\begin{document}

\maketitle

\begin{abstract}
Large Language Models (LLMs) have impressive capabilities in text understanding and zero-shot reasoning. However, delays in knowledge updates may cause them to reason incorrectly or produce harmful results. Knowledge Graphs (KGs) provide rich and reliable contextual information for the reasoning process of LLMs by structurally organizing and connecting a wide range of entities and relations. Existing KG-based LLM reasoning methods only inject KGs' knowledge into prompts in a textual form, ignoring its structural information. Moreover, they mostly rely on close-source models or open-source models with large parameters, which poses challenges to high resource consumption. To address this, we propose a novel \textbf{Light}weight and efficient \textbf{P}rompt learning-\textbf{R}eas\textbf{O}ning \textbf{F}ramework for KGQA (LightPROF), which leverages the full potential of LLMs to tackle complex reasoning tasks in a parameter-efficient manner. Specifically, LightPROF follows a “Retrieve-Embed-Reason” process, first accurately, and stably retrieving the corresponding reasoning graph from the KG through retrieval module. Next, through a Transformer-based Knowledge Adapter, it finely extracts and integrates factual and structural information from the KG, then maps this information to the LLM’s token embedding space, creating an LLM-friendly prompt to be used by the LLM for the final reasoning. Additionally, LightPROF only requires training Knowledge Adapter and can be compatible with any open-source LLM. Extensive experiments on two public KGQA benchmarks demonstrate that LightPROF achieves superior performance with small-scale LLMs. Furthermore, LightPROF shows significant advantages in terms of input token count and reasoning time.
\end{abstract}

%

\section{Introduction}
With the emergence of more Large Language Models (LLMs), their continuously improving performance has brought substantial innovations to the field of Natural Language Processing (NLP) \cite{llmsurvey,llama,gpt4,gemini,chatglm}. The ``\textit{emergent abilities}" displayed under extensive training data and vast parameters allow LLMs to excel in complex zero-shot tasks \cite{emergent}. Despite their effectiveness, LLMs often struggle with knowledge-intensive tasks due to limited task-specific prior knowledge and understanding capabilities \cite{Tog}. Additionally, the costly and time-consuming training process of LLMs presents considerable challenges in continuously updating and maintaining their knowledge bases.

To address the aforementioned challenges, it is crucial to enable LLMs to access a reliable and continuously updated knowledge base to support more accurate and interpretable reasoning \cite{unifying}. Knowledge Graphs (KGs) are ideally suited for this purpose, as they offer a structured semantic framework that delivers both accessible and timely information. 
Knowledge Graph Question Answering (KGQA), as a common knowledge-intensive task, existing work has explored methods for integrating LLMs with KGs to conduct KGQA reasoning \cite{structgpt,retrieve,kaping,wen2023mindmap,Tog,knowledgenavigator}. Broadly speaking, current KG-empowered LLM reasoning primarily involves retrieving information from KGs and incorporating the results into LLM input prompts, leveraging the LLMs' reasoning capabilities to address questions. 

While LLMs reasoning on KGs holds great promise, several challenges remain: 
Firstly, the content of KGs is often represented directly as extensive textual content, which fails to effectively convey the rich logical relationships within their graph structure that are crucial for reasoning. In previous work, the content of KGs was presented in input prompts as multidimensional lists or in natural language form, making it difficult to clearly express the complex relationships and hierarchical structures within them.
Secondly, retrieval and reasoning on KGs demand a high number of LLM calls and substantial LLM reasoning power. Previous work used an iterative approach starting from the question entity, gradually obtaining information for reasoning. This increased the number of LLM calls, sacrificed reasoning efficiency, and diminished feasibility. The textual content describing KGs is vast, requiring not only a larger context window but also a more powerful LLM to ensure that no information is missed while avoiding the generation of incorrect answers in the redundant context.

In response to these challenges, we propose a Retrieve-Embed-Reason framework for LLMs, which is a novel \textbf{Light}weight and efficient \textbf{P}rompt learning-\textbf{R}eas\textbf{O}ning \textbf{F}ramework called \textbf{LightPROF}, designed to provide small-scale LLMs with stable retrieval and efficient reasoning capabilities. 
The framework is structured around three core components: the Retrieval, Embedding, and Reasoning modules. The Retrieval module utilizes relation as the fundamental retrieval unit and limits the retrieval scope based on the question’s semantics to obtain the reasoning graph needed to answer the question. This approach not only boosts the accuracy and stability of retrieval but also considerably narrows the search space and reduces the need for frequent LLM invocations. Next, the Embedding module introduces a small and refined Transformer-based Knowledge Adapter that extracts and integrates the textual and structural information from the reasoning graph, generating representations perfectly suited for the LLM. This module offers an efficient and streamlined way of encoding information, addressing potential ambiguity and information redundancy while reducing the required input token count and context window size, resulting in a more accurate and efficient reasoning process. Finally, The Reasoning module combines the embedded representation vectors with carefully designed natural language prompts, allowing the LLM to derive the final answer. This design allows LightPROF to seamlessly support any open-source LLM and various KGs, requiring only the tuning of the Knowledge Adapter during training, without needing to update the costly and time-consuming LLM.
Our contributions are summarized as follows:
\begin{itemize}
	\item To the best of our knowledge, it is the first framework that transforms both the textual content and graph structure of KGs into embeddings used to prompt LLMs.
	\item We propose LightPROF, a lightweight and efficient prompt-learning reasoning framework that provides small-scale LLMs with stable retrieval and efficient reasoning capabilities, requiring far fewer training parameters compared to the LLM itself.
	\item Extensive experiments conducted on two KGQA datasets demonstrate the superiority of our proposed LightPROF, surpassing methods that use large-scale LLMs (such as LLaMa-2-70B, ChatGPT). Further analysis shows that LightPROF has significant efficiency advantages in terms of input token count and reasoning time.
\end{itemize}

\section{Related Work}
\subsubsection{LLM Prompt Engineering.}In expanding the capabilities of LLMs, prompt engineering has become a crucial technology. It maximizes the performance of LLMs across different applications and research domains by designing special task instructions (i.e., prompts) without altering model parameters \cite{systematic,Saravia_Prompt_Engineering_Guide_2022}. Many studies have been proposed on prompt engineering, spanning from zero-shot prompts \cite{gpt2} and few-shot prompts \cite{gpt3} to Chain-of-Thought (CoT) \cite{cot} and its derivatives such as Tree-of-Thoughts (ToT) \cite{tot,tot2} and Graph-of-Thoughts (GoT) \cite{got}. Additionally, to address the issues of poor robustness and weak expressiveness in discrete prompts, many studies have explored soft prompts \cite{prefix,structured,llaga,letyour}, demonstrating their effectiveness and feasibility in various NLP tasks and structured data representations. Proficiency in prompt engineering can enhance the understanding of the strengths and weaknesses of LLMs.
\subsubsection{KG-based LLM Reasoning.}KGs store a vast amount of explicit and structured knowledge that can effectively enhance the knowledge awareness of LLMs \cite{unifying}. Therefore, researchs have been conducted on using KGs to enhance LLMs’ pre-training and generation techniques. Compared to natural language, KGs have clearer structured logic, which can better guide reasoning. Many studies use factual triples from KGs to construct corpora and employ various pre-training tasks to enhance the capabilities of LLMs \cite{siren,revisit,yu2022jaket,sun2021ernie}. However, this approach causes KGs to lose their advantages of interpretability and dynamism, and may also face catastrophic forgetting issues during the training process \cite{hu2023survey}. 

Therefore, constructing LLM prompts using factual information from KGs is a more flexible, convenient, and secure solution, and our method belongs to this kind of approach. For example, 
KAPING \cite{kaping} retrieves factual knowledge from KGs based on the semantic similarity of the question, adds it to the question as a prompt, and then uses the LLM to generate answers.
KG-GPT \cite{Kg-gpt} uses LLMs to perform reasoning on KG data through three steps: sentence segmentation, graph inference, and reasoning. StructGPT \cite{structgpt} constructs an specialized interface for KG and proposed an Iterative Reading and Reasoning (IRR) framework for LLMs to solve KG-based tasks using this interface. ToG \cite{Tog} utilizes LLMs to iteratively perform beam search on KGs, discovering reasoning paths and returning the most probable reasoning results. KnowledgeNavigator \cite{knowledgenavigator} enhances LLM reasoning by more efficiently and accurately retrieving external knowledge from KGs. While the aforementioned methods have demonstrated commendable performance, they uniformly represent KGs in natural language, which can introduce information redundancy and confusion, ultimately leading to incorrect reasoning.


\section{Preliminaries}

\textbf{Knowledge Graph (KG)} is a data structure that stores a vast quantity of knowledge in the form of triples: $\mathcal{G}=\{(h,r,t)|h,t\in\mathcal{E},r\in\mathcal{R}\}$,where $\mathcal{E}$ and $\mathcal{R}$ denote the set of entities and relations, respectively. A triple $\langle h,r,t\rangle $ represents the existence of a relation $r$ between the head entity $h$ and the tail entity $t$.\\
\textbf{Anchor Entities} are a set of entities: $B = \{b_1,b_2,\ldots,b_K\}$ that are referenced in the KG-based question, where $b_k\in\mathcal{E}$ denotes the $k$-th entity in the question $q$.\\
\textbf{Relation Link} is a sequence of relations: $l = \{r_1,r_2,\ldots,r_J\}$, initiated by an anchor entity for J hop exploration, where $r_j\in\mathcal{R}$ denotes the $j$-th relation in the relation link.\\
\textbf{Reasoning Path} represents a concrete example of the relation link $l$ within the KG of anchor entity  $b_1\in B$: $R_l = \{b_1,r_1,e_1,r_2,\ldots, r_M,e_M\}$, where $r_m\in l$ and $e_m\in\mathcal{E}$ denote the $m$-th   relation and entity in $R_l$, respectively.

\begin{figure}[t]
\centering
\includegraphics[width=1\columnwidth]{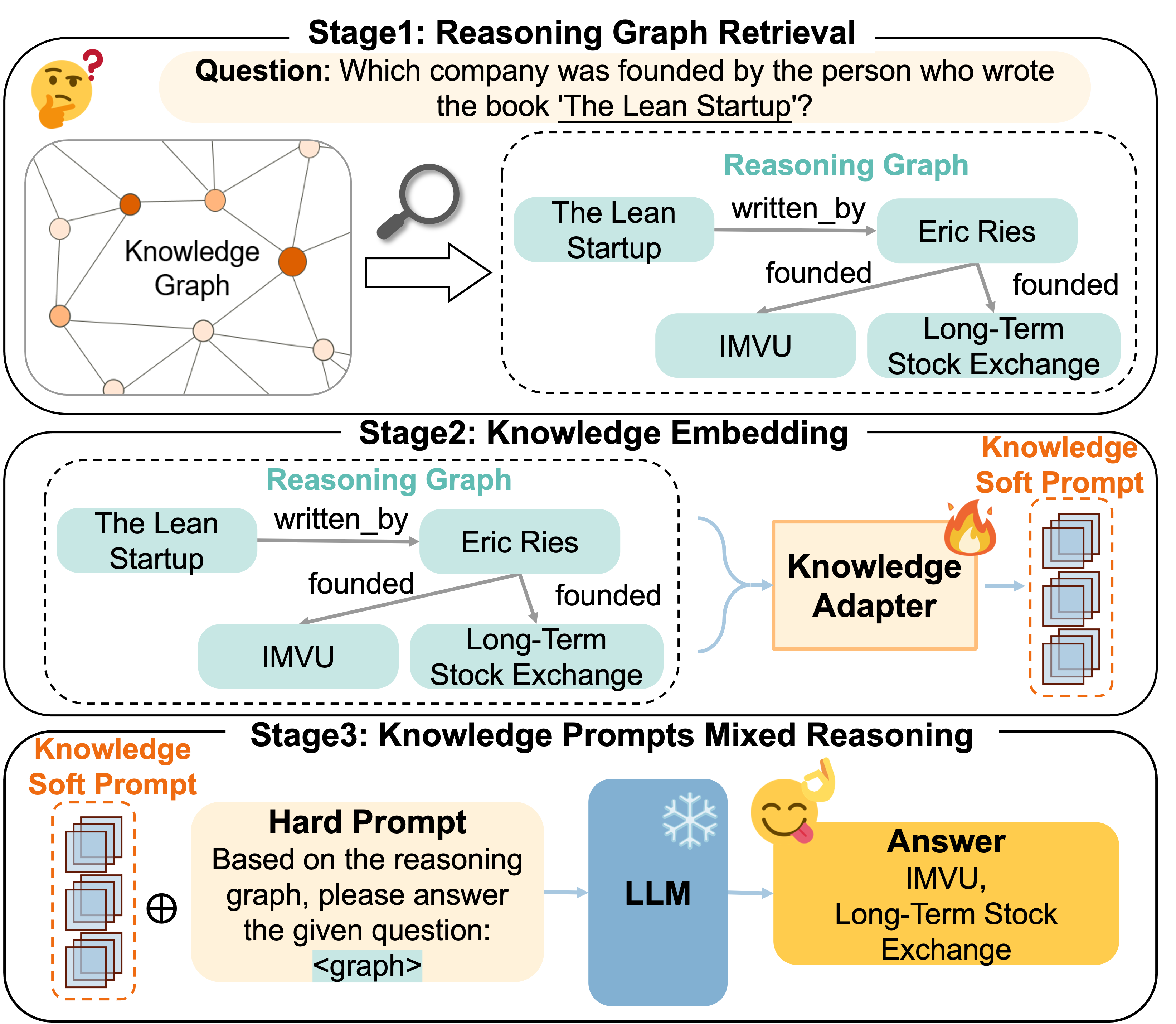} 
\caption{The architecture of our proposed Retrieve-Embed-Reason framework for knowledge graph question answer. }
\label{Fig.main1}
\end{figure}

\section{Methodology}
We design the LightPROF framework, which achieves efficient complex KG problem reasoning under small-scale LLMs through precise retrieval and fine-grained structured data processing capabilities. As shown in Figure~\ref{Fig.main1}, our proposed Retrieve-Embed-Reason framework contains three stages: \textbf{Reasoning Graph Retrieval}, \textbf{Knowledge Embedding}, and \textbf{Knowledge Prompts Mixed Reasoning}.
\subsection{Stage1: Reasoning Graph Retrieval}
\begin{figure}[t]
\centering
\includegraphics[width=1\columnwidth]{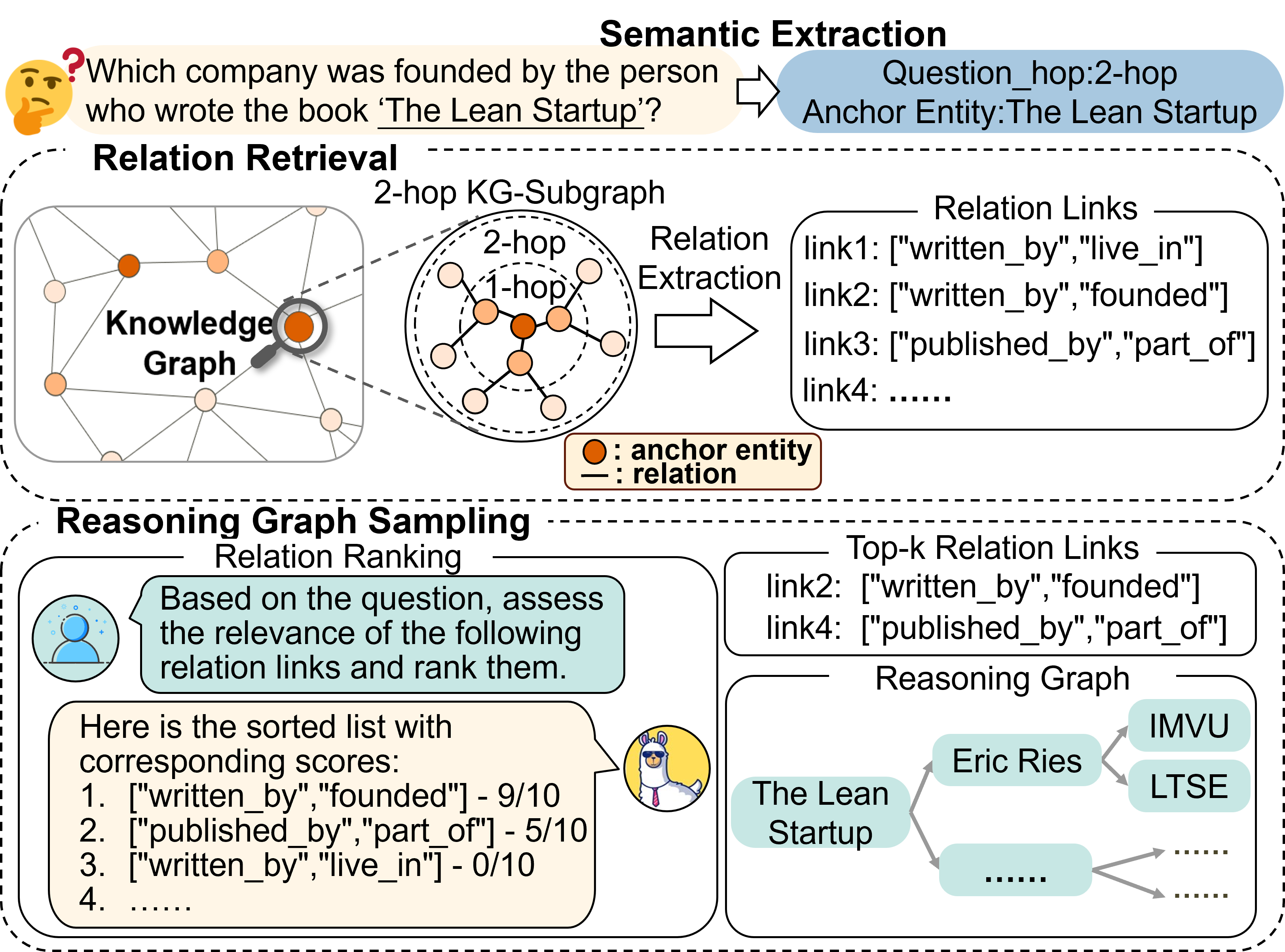} 
\caption{Three Steps Retrieval Module, including: semantic extraction, relation retrieval, and reasoning graph sampling.}
\label{Fig.detail1}
\end{figure}

For the complex multi-hop KGQA task, the question \textit{``How to efficiently, accurately, and stably retrieve information from a KG based on a question?''} is paramount. To address this critical issue, we devide the retrieval module into three steps: semantic extraction, relation retrieval, and reasoning graph sampling, as depicted in Figure~\ref{Fig.detail1}.
\subsubsection{Semantic Extraction.}

For a given question $q$, our goal is to extract relevant semantics (i.e., the number of hops $h_q$ and anchor entities $B$) from the KG to narrow the retrieval scope while preserving the essential reasoning knowledge. This approach enables the retrieval and construction of a highly relevant and precise reasoning graph \cite{knowledgenavigator}.  Specifically, we fine-tune a pre-trained language model (PLM), such as BERT, to learn the number of hops $h_q$ in KG required for reasoning, based on the semantic vector $V_q$ of the query $q$. $H$ is the maximum number of hops in the dataset, which can be framed as a classification task:
\begin{equation}    V_q = \operatorname{PLM}(q)\end{equation}
\begin{equation}h_q=\arg\max_hP(h|V_q),h = 1,2,\ldots,H. \end{equation}

\subsubsection{Relation Retrieval.} Relations in KGs describe the specific connections between two entities, providing semantic clarity for their interactions and substantially enriching the information content of KGs. Many studies currently utilize semantically rich relation links for KG reasoning tasks \cite{DBLP:conf/emnlp/XiongHW17,ijcai2022p325,DONG2023101900}. More crucially, relations in KGs demonstrate more stability and intuitiveness compared to the continuously changing and complex entities \cite{ijcai2023p734}. To gather as much relevant knowledge as possible, we adopt a search for relation links in the KG based on anchor entities $B$ and the predicted hop $h_q$. Specifically, the model first selects an anchor entity and then employs a constrained breadth-first search (BFS) with a depth limit of $h_q$. This process is designed to collect all relation links originating from the anchor entity  $B$  and extending up to a predetermined length of $h_q$.
\subsubsection{Reasoning Graph Sampling.} First, the retrieved relation links are fed into a LLM. Subsequently, the LLM calculates scores and ranks them according to their semantic relevance to the question $q$. Then, we select the top-$k$ relevant links. Finally, we sample in KG based on the selected relation links, extracting multiple reasoning paths $\{R_1,R_2,\ldots,R_N\}$ to construct a refined reasoning graph, denoted as $G_R$.
\subsection{Stage2: Knowledge Embedding}
KGs typically encompass a rich array of complex structural information, including subgraph structures, relational patterns, and the relative relation between entities \cite{DBLP:journals/corr/abs-2310-06671}. Such structural information is essential for LLMs to gain a deep understanding of KGs. However, the natural language expression of KG structural information contains redundancy and confusion, which cannot directly reveal its inherent nature, thus impeding LLMs from effectively utilizing this information.
\begin{figure}[t]
\centering
\includegraphics[width=0.8\columnwidth]{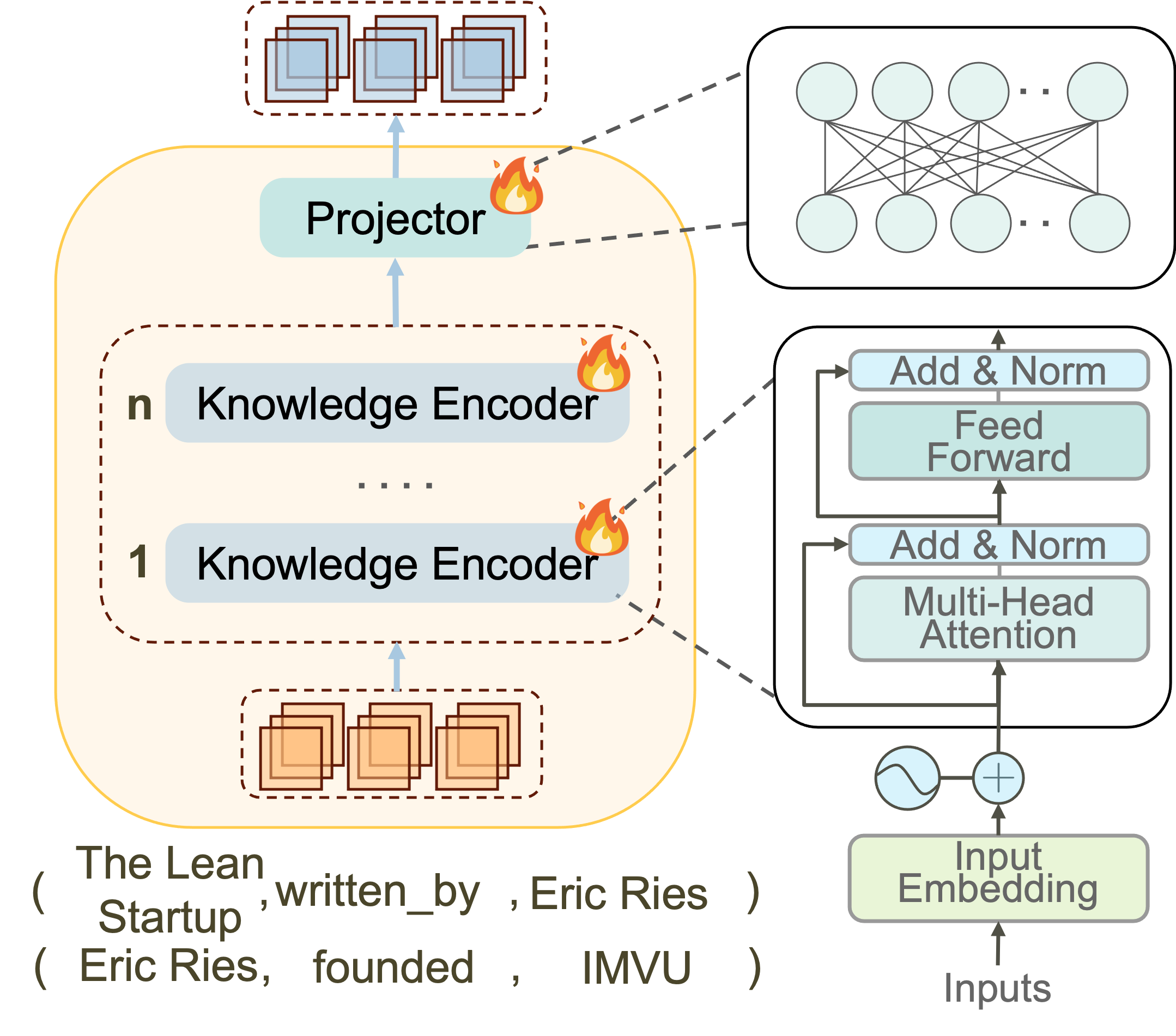} 
\caption{Illustration of the Knowledge Adapter and the schematic representation of its crucial components.}
\label{Fig.detail2}
\end{figure}

To address the aforementioned challenge, as inspired by \cite{llaga,letyour}, we propose a refined and compact Knowledge Adapter that can encode textual information in the reasoning graph while extracting its structural information, as illustrated in Figure~\ref{Fig.detail2}. By combining textual information with structural details at a fine granularity, Knowledge Adapter aids the model in deeply comprehending the knowledge within the reasoning graph, enabling more precise reasoning.

Specifically, we assume that the reasoning graph $G_R=\{R_n\}_{n=1}^N$ is composed of $N$ reasoning paths, each of which is decomposed into a set of triples ${\mathcal{T}}^{n}=\{(h_{i}^{n},r_{i}^{n},t_{i}^{n})| i\in[1,h_q]\}$, where $h_q$ is the number of reasoning hops.
Subsequently, $\operatorname{Embed}(\cdot)$, i.e., BERT, is used to obtain the relational embedding $e_i^{r}$ for each triple:
\begin{equation}
\mathbf{e}_i^{r}=\operatorname{Embed}(r_{i}^{n}).
\end{equation}
We can obtain the entity embeddings $e_i^{h}, e_i^{t}$ in the same way.
Next, we aim to capture both the local and global interactions between each entity and relation. We first use $\operatorname{StructEmb}(\cdot)$ to encode the local structural information $\mathbf{s}_i$ of $i$-th triple in ${\mathcal{T}}^{n}$. Then, a linear layer $\operatorname{Linear}(\cdot)$ is used to aggregate the global structural information $\mathbf{z}^{s}$ from the entire reasoning path $R_n$:

 \begin{equation}
 \begin{split}
    \mathbf{s}_i=\operatorname{StructEmb}(\mathbf{e}_i^{h},\mathbf{e}_i^{r},\mathbf{e}_i^{t}),\\
    \mathbf{z}^{s} = \operatorname{Linear}(\mathbf{s}_1,\mathbf{s}_2,\ldots,\mathbf{s}_{h_q}).
\end{split}
\end{equation}


Additionally, to capture the textual information of the reasoning path $R_n$, we use $\operatorname{Fusion}(\cdot)$  to combine the text-level information of all entities and relations in $R_n$. We first obtain the combined text representation  $\mathbf{z}^{t_{h}}$ of all head entities as follows:
\begin{equation}\begin{aligned}
\mathbf{z}^{t_{h}}&=\operatorname{Fusion}(\mathbf{e}_1^{h},\ldots,\mathbf{e}_{h_q}^{h}).
\end{aligned}\end{equation}
Then, the combined text representations of relations $\mathbf{z}^{t_{r}}$ and tail entities $\mathbf{z}^{t_{t}}$ can be obtained in the same way.
Afterwards, these vectors are consolidated into a single vector $\mathbf{z}^{t}$ to represent the comprehensive textual information of the entire reasoning path $R_n$:

\begin{equation}\mathbf{z}^{t}=f_c(\mathbf{z}^{t_{h}},\mathbf{z}^{t_{r}},\mathbf{z}^{t_{t}}), \end{equation}
where $f_c(\cdot)$ is the consolidation function. While $f_c(\cdot)$ can be complex neural networks or language models, to preserve the semantic integrity of the text and reduce the model’s training complexity, we use a simple concatenation operation to form a composite vector that encapsulates all the textual information of the entire reasoning path.

Finally,  we use $\operatorname{KnowledgeEncoder}(\cdot)$ to seamlessly integrate the obtained comprehensive textual information $\mathbf{z}^{t}$ and global structural information $\mathbf{z}^{s}$, deriving a fused representation of the reasoning path,  as shown in Figure~\ref{Fig.detail2}:
\begin{equation}\mathbf{z}^f=\operatorname{KnowledgeEncoder}([\mathbf{z}^{t},\mathbf{z}^{s}])\end{equation}

In this way, the Knowledge Encoder can effectively encode each reasoning path in the reasoning graph into a single token, significantly improving the token utilisation efficiency of the LLM and enhancing the representational capacity of the reasoning paths. During the encoding process, the Knowledge Encoder captures not only rich textual information from the reasoning graph but also crucial structural information. Since the fused information $\mathbf{z}^f$ contains both textual and structural elements, the model can more fully understand the meaning embedded in each reasoning path during inference. This multidimensional information representation enhances the model’s sensitivity to context , facilitating more effective deep semantic analysis and reasoning. Consequently, this information integration allows the model to more accurately capture the complex interactions between semantics and structure, thereby enhancing the accuracy and depth of reasoning.

By aggregating all paths $\{R_n\}_{n=1}^N$, we obtain the representational sequence $[\mathbf{z}_1^f,\mathbf{z}_2^f,\ldots,\mathbf{z}_N^f]$ of the reasoning graph $G_R$. Before inputting the sequence into the LLM, a dimension transformation is necessary.  Due to the differences between the embedding space of the Knowledge Encoder and the input space of the LLM, directly using these tokens would be ineffective. Therefore, we develop a trainable projector $\Phi(\cdot)$,  which maps these tokens into the token embedding space of the LLM. As a result, this process generates an input sequence suitable for the LLM, which we refer to as the knowledge soft prompt $p_{\rm{s}}$:
\begin{equation}p_{\rm{s}}=\Phi([\mathbf{z}_1^f,\mathbf{z}_2^f,\ldots,\mathbf{z}_N^f]).\end{equation}
Here we set  $\Phi(\cdot)$ as a two-layer multilayer perceptron. 
Following the aforementioned process, the Knowledge Adapter is able to encode the textual representation of the reasoning graph into the corresponding knowledge soft prompt. Importantly, all parameters of this adapter are derived from the parameters of the Knowledge Encoder and Projector, which are the only components requiring tuning during the LightPROF training process.
\subsection{Stage3: Knowledge Prompts Mixed Reasoning}
LLMs have acquired extensive knowledge through broad training on large corpora. However, despite their proficiency in general knowledge, LLMs show notable deficiencies in processing specialized knowledge, complex long logic chains, and multi-hop knowledge reasoning, which mainly stem from the limitations of their pre-training data. Additionally, although the knowledge base of LLMs can be expanded through retraining, this method is usually costly and time-consuming \cite{Tog}. More seriously, retraining may lead to catastrophic forgetting of existing knowledge in the model \cite{zhang2024graphtranslator}. Thus, this presents certain challenges in keeping LLMs' knowledge up-to-date.
To avoid the aforementioned challenges, we freeze the parameters of the LLM during the LightPROF training process and use a combination of soft prompts and hard prompts to guide the model to answer questions more precisely and efficiently, which can be seen in Figure~\ref{Fig.main1}. 

Specifically, the input to the LLM is organized in a chat format, where instructions and questions are combined using carefully designed text templates, which we call hard prompts. During the encoding phase of the LLM, we insert the knowledge soft prompt, representing the reasoning graph, into specific locations of the hard prompt to effectively inject external knowledge, as shown in Figure~\ref{Fig.main1}. This approach allows the LLM to autonomously and accurately answer questions based on the given input content without the need for parameter updates. By this method, we not only maintain the stability of the model but also enhance its performance and efficiency within specific knowledge domains. 

The training objective of LightPROF is to maximize the likelihood of generating correct answers $\mathcal{A}$ for all samples in the dataset $\mathcal{D}$. This can be compatible with the task of next-token prediction, a fundamental method for training generative models. The training goal can be articulated as:
\begin{equation}
    \arg\max_{\mathcal{A}}P_{\rm{llm}}(\mathcal{A}| p_{\rm{p}})=\sum^{\mathcal{D}}\sum_{t=1}^{|\mathcal{A}|}\log P_{\rm{llm}}(a_t|a_{1:t-1},p_{\rm{h}},p_{\rm{s}}),
\end{equation}
where $p_{\rm{p}}$ is the input sequence that includes both hard prompt $p_{\rm{h}}$ and soft prompt $p_{\rm{s}}$, and $a_t(t=1,2,\ldots,|\mathcal{A}|)$ is the $t$-th token of the output sequence. Notably, when $t=1$, $a_{1:t-1}$ is the model’s beginning-of-sequence (BOS) token.
\section{Experiments}
In this experiment, we will thoroughly discuss the following questions. \textbf{Q1}: How significantly can LightPROF enhance LLMs' performance in KGQA tasks? \textbf{Q2}: Can LightPROF be integrated with different LLM backbones to enhance performance? \textbf{Q3}: Can LightPROF achieve efficient input and stable output with small-scale LLMs?
\subsection{Datasets}
We train and evaluate LightPROF's multi-hop reasoning capabilities on two public datasets based on the Freebase knowledge graph \cite{Freebase}: WebQuestionsSP(WebQSP) \cite{qsp} and ComplexWebQuestions(CWQ) \cite{cwq}. Based on previous works, we utilize match accuracy (Hits@1) to evaluate whether the model's top-1 answer is correct.

\begin{itemize}
    \item \textbf{WebQSP} is a benchmark with fewer questions but a larger knowledge graph, consisting of 4,737 questions. Each question includes a topic entity, a reasoning chain, and a SPARQL query to find the answer. The answer entity requires up to 2-hop reasoning on the Freebase.
    \item \textbf{CWQ} is a benchmark specifically designed for complex knowledge graph question answering research. It includes 34,689 question-answer pairs, built upon the WebQSP dataset. It involves automatically creating more complex SPARQL queries and generating corresponding natural language questions, thereby creating a wide and diverse range of question types. These questions require up to 4-hop reasoning on Freebase.
\end{itemize}

\subsection{Baselines}
We consider three types of baseline methods: full fine-tuning methods, vanilla LLM methods, and LLM+KGs methods. The full fine-tuning methods include KV-Mem \cite{KV-Mem}, EmbedKGQA \cite{EmbedKGQA}, TransferNet \cite{shi2021transfernet}, NSM \cite{NSM}, KGT5 \cite{KGT5}, GraftNet \cite{GraftNet}, PullNet \cite{pullnet}, UniKGQA \cite{unikgqa}. Vanilla LLM methods include LLaMa series models \cite{llama}. LLM+KGs methods include StructGPT \cite{structgpt}, ToG \cite{Tog}, KnowledgeNavigator \cite{knowledgenavigator}, AgentBench \cite{agentbench}. Notably, to ensure fair comparisons, the LLM+KGs methods we select do not involve fine-tuning the LLMs, i.e., all of them are zero-shot methods without any training of the LLM.
\subsection{Implementation}
To demonstrate the plug-and-play convenience and parameter efficiency of LightPROF, we conduct experiments on two small-scale 
language models in the LLaMa series: LLaMa-7B-chat \cite{llama} and LLaMa-8B-Instruct\footnote{https://ai.meta.com/blog/meta-llama-3/}. The model was optimized over one training epoch with a batch size of 4. The initial learning rate was set at 2e-3, adjusted using a cosine annealing schedule to enhance the model's learning efficiency during training. All experiments are conducted using the PyTorch toolkit on NVIDIA A800 GPU.

The Knowledge Encoder module is based on the BERT model. The module includes a two-layer MLP Projector that maps dimensions to the LLM's input dimension.
\subsection{Q1: Performance Comparison}
\subsubsection{Main Result.}We evaluate LightPROF against three categories of baseline methods: full fine-tuning, vanilla LLM, and LLM+KGs approaches. As illustrated in Table \ref{tab.main}, LightPROF not only excels in simple questions but also demonstrates high performance in scenarios requiring deep reasoning and complex query handling. Specifically, LightPROF significantly surpasses the state-of-the-art model on the WebQSP dataset (83.7\% vs. 75.1\%) and also excels on the more complex CWQ dataset (59.3\% vs. 57.6\%). These outcomes validate our framework’s excellent capability in addressing KGQA tasks, emphasizing LightPROF’s efficacy in managing multi-hop and complex challenges.

Compared to vanilla LLMs and LLM+KGs methods that utilize plain text prompts, LightPROF’s significant improvement indicates that soft prompts produced by the Knowledge Adapter can effectively encapsulate more complex structural knowledge than discrete text, being concise, informative, and highly expressive, thus enhancing LLM’s understanding of KG information. It is noteworthy that our framework outperforms other large-scale models in all experimental conditions. For example, our framework excels, particularly in reasoning through complex problems, compared to ToG \cite{Tog} with LLaMa2-70B-Chat and StructGPT \cite{structgpt} with ChatGPT. Additionally, even with the smaller LLaMa2-7b version, our framework competes effectively with other large-scale models, underscoring the efficiency and optimization of our framework’s design.
\begin{table}[!ht]
    \centering
    \begin{tabular}{p{5cm}cc}
    \toprule
        Methods & WebQSP & CWQ \\ \midrule
        KV-Mem & 46.7 & 18.4  \\ 
        EmbedKGQA & 66.6 & 45.9  \\ 
        NSM & 68.7 & 47.6  \\ 
        KGT5 & 56.1 & 36.5  \\ 
        GraftNet & 66.4 & - \\ 
        PullNet & 68.1 & - \\ 
        TransferNet & 71.4 & 48.6  \\ 
        UniKGQA & \underline{75.1} & 50.7  \\ \midrule
        LLaMa2-7B-Chat& 61.4 & 31.5 \\
        LLaMa2-70B-Chat & 57.4 & 39.1 \\
        \midrule
        ToG (LLaMa2-70B) & 68.9 & \underline{57.6}  \\ 
        StructGPT (ChatGPT) & 72.6 & 54.3 \\ 
        AgentBench & 47.8 & 24.8 \\
        KnowledgeNavigator(LLaMa2-70B) & 71.8 & - \\ 
        \midrule
        \textbf{LightPROF (LLaMa3-8B)} & \textbf{83.8} & \textbf{59.3} \\ 
        \textbf{LightPROF (LLaMa2-7B)} & 71.2 & 48.5 \\ 
        \bottomrule
    \end{tabular}
    \caption{Performance comparison of LightPROF with baselines on the two datasets. Bold and underlined typefaces indicate optimal and sub-optimal methods, respectively.}
    \label{tab.main}
\end{table}
\subsubsection{Ablation Study.} 
\begin{table}[!ht]
    \centering
    \begin{tabular}{p{5cm}cc}
    \toprule
        Methods & WebQSP & CWQ  \\ \midrule
        LightPROF & \textbf{83.77} & \textbf{59.26}  \\ \midrule
        LightPROF w/o Struct & 82.36 & 58.05  \\ 
        LightPROF w/o Train & 80.37 & 55.63  \\ \midrule
        LightPROF w/ Random Retrieve & 53.44 & 46.84  \\ \bottomrule
    \end{tabular}
    \caption{Model ablation study of our LightPROF framework.}
    \label{tab.ablation}
\end{table}
An ablation study is performed on LightPROF to investigate the specific effects of the Knowledge Adapter on KGQA task performance. We examine three variants: (1) \textit{w/o Struct}, removing the structural information included in the knowledge embedding process, (2) \textit{w/o Train}, without training the Knowledge Encoder, and (3) \textit{w/ Random Retrieve}, randomly retrieve reasoning paths from KGs. The results are displayed in Table \ref{tab.ablation}.

The results indicate that the integration of structural information is crucial for the model's understanding and handling of entities and relationships in complex queries. The incorporation of structural information significantly enhances the model's utilization efficiency of data in the knowledge graph. Continuous training of the Knowledge Encoder is also essential for enhancing the model's comprehension and generation of knowledge representations. This training process notably improves the model's capability to encode complex structural knowledge, allowing it to more accurately respond to queries rooted in deep knowledge. Moreover, randomly retrieved reasoning paths can cause significant damage to performance, highlighting the importance of an accurate and stable retrieval module.

Additionally, we explore different structural encoders. The structural encoder used in our framework encodes triples as Head (H) + Relation (R) - Tail (T). Results in Table \ref{tab.struct} show that the performance of the H+R+T encoding method slightly declines due to its inability to distinguish the order of the triples, \emph{e.g.}, the structural information derived from (Eric Ries, founded, IMVU) and (IMVU, founded, Eric Ries) is identical, reducing the model's capacity to understand structural information. In contrast, LightPROF can better capture structural information within the reasoning graph and integrate it at a finer granularity, enhancing the model's understanding, particularly in scenarios involving complex structured data reasoning.
\begin{table}[!ht]
    \centering
    \begin{tabular}{p{5cm}cc}
    \toprule
        Methods & WebQSP & CWQ  \\ \midrule
        LightPROF(H+R+T) & 83.68 & 58.32  \\ 
        LightPROF(H+R-T) & \textbf{83.77} & \textbf{59.26}  \\ \bottomrule
    \end{tabular}
    \caption{Performance impact of different structure encoder in LightPROF.}
    \label{tab.struct}
\end{table}

\begin{table*}[!ht]
    \centering
    \begin{tabular}{ll}
    \toprule
        \textbf{Question} & what drugs lindsay lohan abuse?  \\ \midrule
        \textbf{Answer} & $[$ ``Alcoholic beverage", ``Cocaine"$]$   \\ \midrule
        \textbf{StructGPT} & \makecell[l]{The relevant relation: celebrities.celebrity.substance\_abuse\_problems\\ 
        The possible constraints: celebrities.substance\_abuse\_problem.substance: Alcoholic beverage\\ 
        The final answers: Alcoholic beverage}  \\ \midrule
        \textbf{LightPROF} & \makecell[l]{Number of Hops: 2\\
        Relation Links:\\
        $[$‘base.popstra.celebrity.substance\_abuse’, ‘base.popstra.substance\_abuse.substance’$]$ - 9/10\\
        $[$‘base.popstra.celebrity.substance\_abuse’, ‘base.popstra.substance\_abuse.abuser’$]$ - 9/10\\
        Based on the knowledge graphs, please answer the given question. Please keep the answer as \\simple as possible and return all the possible answers as a list. knowledge graphs: $<graph>$ \\
        $[$``Cocaine", ``Alcoholic beverage"$]$
        }   \\
    \bottomrule 
    \end{tabular}
    \caption{Case Study of LightPROF and StructGPT on the WebQSP Dataset.}
    \label{tab.case}
\end{table*}

\subsection{Q2: Plug-and-Play}
For our framework, any open-source LLM capable of accepting token embedding inputs is suitable. In this section, we evaluate the effectiveness of integrating different LLMs within LightPROF. As illustrated in Table \ref{tab.plug}, the results demonstrate that the LightPROF framework significantly enhances the performance of integrated LLMs, regardless of the baseline performance of the original models. LightPROF enhances the model's capability to address complex KG questions through effective integration and optimization of structured data. This plug-and-play integration strategy does not require costly fine-tuning of LLMs, making it particularly suitable for quickly enhancing existing models' performance on KGQA task.
\begin{table}[!ht]
    \centering
    \begin{tabular}{p{5cm}cc}
    \toprule
        Methods & WebQSP & CWQ  \\ \midrule
        Llama2-7b & 61.36 & 31.49  \\ 
        LightPROF (Llama2-7b) & 71.19 & {48.48}  \\ 
        Llama3-8b & 66.83 & 48.87  \\ 
        LightPROF (Llama3-8b) & \textbf{83.77} & \textbf{59.26}  \\ \bottomrule
    \end{tabular}
    \caption{The performance of integrating various LLMs into the LightPROF framework.}
    \label{tab.plug}
\end{table}
\subsection{Q3: Efficient Input and Stable Output}
\subsubsection{Efficiency Results.} A series of efficiency tests are conducted to compare the performance of LightPROF and StructGPT \cite{structgpt} when processing the WebQSP dataset. Specifically, the models' runtime, the total number of input tokens, and the average \underline{N}umber of tokens \underline{P}er \underline{R}equest (NPR) are measured, with results presented in Table \ref{tab.Efficient}. The table shows that LightPROF is more time-efficient when processing the same dataset, with a 30\% reduction in time cost (1:11:49 vs. 1:42:12). Regarding the total number of input tokens, LightPROF and StructGPT show a significant difference (365,380 vs. 24,750,610), demonstrating that LightPROF is more economical in input processing, reducing token usage by approximately 98\%. Furthermore, LightPROF's NPR value is 224, significantly lower than StructGPT's 6400. This comparison further highlights LightPROF's advantage in the number of tokens needed per request, showcasing its more precise and resource-efficient handling of each request, validating LightPROF's effectiveness when integrating small-scale LLMs.
\begin{table}[!ht]
    \centering
    \begin{tabular}{p{3cm}ccc}
    \toprule
        Methods & TimeCost & TokenUsed & NPR  \\ \midrule
        LightPROF & \textbf{1:11:49} & \textbf{365,380} & \textbf{224}  \\ 
        StructGPT & 1:42:12 & 24,750,610 & 6400  \\ \bottomrule 
    \end{tabular}
    \caption{Efficiency performance of LightPROF and StructGPT on Llama-3-8b. NPR represents the average number of tokens per request.}
    \label{tab.Efficient}
\end{table}

\subsubsection{Case Study.}As shown in Table \ref{tab.case}, we validate LightPROF's efficient input and stable output capabilities when using small-scale LLMs by comparing its performance with StructGPT to answer complex queries about Lindsay Lohan's drug abuse. The results show that LightPROF not only accurately identify and comprehensively answer the query, but also demonstrate deeper reasoning pathways and overall scoring. In contrast, although StructGPT handled the relevant questions, it failed to fully capture all related answers. Interestingly, we found that LightPROF can consistently generate output that includes only the answers and uses fewer input tokens and less reasoning time. This suggests that LightPROF can effectively integrate and precisely output complex information from knowledge graphs, demonstrating its reliability and practicality in efficiently and accurately handling complex KGQA tasks.

\section{Conclusion}
In this paper, we introduce the LightPROF framework, which accurately retrieves and efficiently encodes KGs to enhance LLM reasoning. To effectively narrow the retrieval scope, LightPROF incrementally samples the KG using stable relationships as units. To achieve efficient reasoning on LLMs with fewer parameters, we develop a delicate Knowledge Adapter that can effectively parse graph structures and perform fine-grained information integration, thus condensing the reasoning graph into a smaller number of tokens and achieving comprehensive alignment with the LLM's input space through the Projector. Experimental results show that our framework outperforms other baseline methods, particularly those involving large-scale language models. In comparison to other methods based exclusively on text, our knowledge soft prompts integrate a more comprehensive range of structural and textual information, making them more easily understood by LLMs. In future work, we plan to explore 1) KG encoders with stronger generalization and compatibility, and design an encoder that can be applied to unseen KG data without retraining. 2) A unified cross-modal encoder capable of encoding multimodal KGs.
\section{Acknowledgments}
The research was supported by the National Natural Science Foundation of China (Grant No. U22B2019).

\bibliography{aaai25}
\end{document}